\documentclass[runningheads]{llncs}

\usepackage{url}
\usepackage{bbold}
\usepackage{multirow}
\usepackage{graphicx}
\usepackage{booktabs}
\usepackage{amsfonts}
\usepackage{amsmath}
\usepackage{enumitem}
\usepackage[sort]{cite}
\usepackage[hidelinks]{hyperref}
\usepackage[noend,algo2e,boxruled,linesnumbered]{algorithm2e}
\usepackage[misc]{ifsym}
\usepackage[dvipsnames,table,xcdraw]{xcolor}
\usepackage{amssymb}
\usepackage{pifont}
\usepackage{latexsym}
\usepackage{algorithm}
\usepackage[utf8]{inputenc}
\usepackage{microtype}
\usepackage{booktabs}
\usepackage{longtable}
\usepackage{bbm}
\usepackage{lipsum}
\usepackage{tikz}
\usetikzlibrary {decorations.pathmorphing} 
\usepackage{comment}
\usepackage{pdfpages}
\usepackage{wrapfig}

\newcommand{\quotes}[1]{``#1''}

\usetikzlibrary{arrows.meta}
\tikzstyle{hidden vertex} = [circle, draw=black,style=dashed]
\tikzstyle{vertex} = [circle, fill=black!10]
\tikzstyle{box} = [fill=black!10,draw = black,minimum width=0.7cm, minimum height=0.7cm]
\tikzstyle{fixed vertex} = [circle, draw=red,fill=red!10]
\tikzstyle{edge} = [-latex,thick]
\tikzstyle{o->} = [{Circle[open]}->, thick]
\tikzstyle{o-o} = [{Circle[open]}-{Circle[open]}, thick]
\tikzstyle{<-->} = [<->, thick]
\tikzstyle{redge} = [->,thick, draw=red]
\tikzstyle{uedge} = [thick, draw=black!20]
\tikzstyle{bluedge} = [->,thick, draw=blue]
\tikzstyle{snakedge} = [-stealth,thick, shorten <=1pt,decorate, decoration={snake,amplitude=.3mm}]
\tikzstyle{array}=[draw, fill=green!40, minimum width = 7mm, minimum height = 7mm]

\DeclareMathOperator*{\argmin}{arg\,min}

\newcommand{\orightarrow}{\hbox{$\circ$}\kern-1.5pt\hbox{$\rightarrow$}}
\newcommand{\leftarrowo}{\hbox{$\leftarrow$}\kern-1.5pt\hbox{$\circ$}}
\newcommand{\oarrowo}{\hbox{$\circ$}\kern-1.5pt\hbox{$-$}\kern-1.5pt\hbox{$\circ$}}

\begin{document}
\title{The Importance of Time \\ in Causal Algorithmic Recourse}

\author{Isacco Beretta\inst{1}\orcidID{0000-0002-0463-6810}
    \and \\
    Martina Cinquini\inst{1}\orcidID{0000-0003-3101-3659}\Letter
}
\authorrunning{Beretta and Cinquini}
%

\institute{
    University of Pisa, Italy, \email{\{name.surname\}@phd.unipi.it} 
}

\maketitle              
\begin{abstract}
The application of Algorithmic Recourse in decision-making is a promising field that offers practical solutions to reverse unfavorable decisions. However, the inability of these methods to consider potential dependencies among variables poses a significant challenge due to the assumption of feature independence. Recent advancements have incorporated knowledge of causal dependencies, thereby enhancing the quality of the recommended recourse actions. 
Despite these improvements, the inability to incorporate the temporal dimension remains a significant limitation of these approaches. 
This is particularly problematic as identifying and addressing the root causes of undesired outcomes requires understanding time-dependent relationships between variables. 
In this work, we motivate the need to integrate the temporal dimension into causal algorithmic recourse methods to enhance recommendations' plausibility and reliability. The experimental evaluation highlights the significance of the role of time in this field. 
\end{abstract}
\section{Introduction}
Counterfactual explanations are becoming one of the most promising solutions to explainability in Machine Learning due to their compliance with legal requirements~\cite{wachter2017counterfactual}, their psychological benefit for the individual~\cite{venkatasubramanian2020philosophical}, and their potential to explore ``what-if'' scenarios~\cite{chou2022counterfactuals}. 
A possible circumstance in which such explanations are required is when a credit scoring model denies a loan to an applicant, and the individual desires to understand what should be different to change the outcome of the AI system (i.e., to have the loan accepted). 
Comprehending an unfavorable decision adds new information about the facts, enhancing human trust in automated decision-making systems. Additional advantages from the user perspective could be gained by learning what \textit{actions} take to reach a different outcome. 
A novel research area~\cite{karimi2020survey}, referred to as \textit{Algorithmic Recourse} (AR), aims at suggesting actionable recommendations that should be performed to reverse unfavorable decisions in the future. Among extensive literature, recent work~\cite{barocas2020hidden} highlights that a significant drawback of AR methods is the implicit assumption of examining features as independently manipulable inputs.
Since the individual’s attributes change may have downstream effects on other features, observing and identifying causal mechanisms is crucial in analyzing real-world scenarios to avoid sub-optimal or infeasible actions. From this perspective, Karimi et al.~\cite{karimi2021, karimi2020algorithmic} propose a fundamental reformulation of the recourse problem, incorporating knowledge of causal dependencies into the process of recommending recourse actions. 
The ability to assess 
the causal relationships 
explicitly guarantees plausible counterfactuals~\cite{byrne2019} and improves the user's perception of a decision's quality since it reflects the tendency of human beings to think in terms of cause-effect~\cite{pearl2018book}.


Despite the recent progress in this field, a significant limitation of current methods is their inability to incorporate the temporal dimension. Neglecting the temporal interdependencies between features and actions can result in erroneous identification of the feature that requires modification at a particular moment, leading to ineffective or sub-optimal recourse actions. As a result, there is a need to devise causal AR techniques that can incorporate temporal information to provide explanations that precisely reflect the complex dynamics of the system and to guarantee that the recommendations offered are reliable and plausible.

This work investigates the usefulness of integrating the temporal dimension into a causal AR problem by incorporating the topological information of the causal graph in the cost function evaluation. Besides, it discusses the necessity of interpreting the causal model as a representation of a dynamical process i.e., one that involves the evolution of its instances over time.

The rest of the paper is organized as follows. 
Section~\ref{sec:related} describes the state-of-the-art related to causal AR. Section~\ref{sec:setting} recalls basic notions for understanding our proposal. Section~\ref{sec:motivation} motivates for our proposal by presenting a brief methodological discussion and a practical example.
The cost function evaluation is presented in Section~\ref{sec:role_time}, while Section~\ref{sec:experiment} reports the experimental results.
Section~\ref{sec:findings} examines the impact of findings on the progress of the XAI field,
Section~\ref{sec:conclusions} summarizes our contributions and discusses open research directions.

\section{Related Works}\label{sec:related}
Most of the existing approaches in the AR literature~\cite{joshi2019towards, karimi2020survey, sharma2019certifai} derive recourse actions through solving an optimization problem that minimizes changes to the individual's feature vector, subject to various plausibility, diversity, and sparsity constraints.
In~\cite{karimi2021} is presented a paradigm shift from recourse via nearest counterfactual explanations. The objective is to find recourse through minimal interventions attempting to use a causal probabilistic framework grounded on Structural Causal Models (SCMs) that fit in the class of additive noise models. Specifically, to seek the minimal cost set of actions in the form of structural interventions that would favorably change the prediction if acted upon, authors exploit structural counterfactual computed deterministically in closed-form by applying the Abduction-Action-Prediction procedure proposed in~\cite{pearl2009causality}. 
A significant drawback of this formulation is the extraction of SCM from the observed data. Indeed, assuming the knowledge of the true causal graph and the structural equations is very challenging and, in some cases, could be unrealistic~\cite{peters2017elements}.  
Therefore, in~\cite{karimi2020algorithmic} is presented two probabilistic approaches that relax such assumption. In both cases, authors suppose the knowledge of the causal graph a priori or postulated by an expert. The first method, referred to \textit{individualized recourse via GP-SCMs}, 
consists of using additive Gaussian noise and Bayesian model averaging to estimate the counterfactual distribution. 
The second approach, also known as \textit{subpopulation-based recourse via CATE}, 
removes any assumptions on the structural equations by computing the conditional average treatment effect of an intervention on individuals similar to the factual subject. 

Recently, there has been a growing interest in extending the formulation of actions and their consequences to incorporate them into a sequential context. This is due to the fact that, in reality, most changes do not occur instantaneously but are part of a process. For instance, in~\cite{naumann21}, the authors propose a model-agnostic method for generating sequential counterfactuals that have the ability to discover multiple optimal solution sequences of varying sequence lengths. 

Furthermore, another novel research direction related to AR entails distinguishing the factors that influence the change in model prediction (i.e., acceptance) from those that contribute to the state of the real world (i.e., improvement)~\cite{barocas2020hidden}. In~\cite{konig2022} is tackled this subject by introducing the first approach that specifically focuses on promoting improvement rather than mere acceptance.

However, to the best of our knowledge, no state-of-the-art methods account for the \textit{temporal relationship} between features and actions. 
This work aims to fill such a gap by incorporating this crucial dimension, which in turn enables us to provide more precise recommendations that reflect the reality of decision-making processes. 
Specifically, our approach evaluates the cost of an action taken by a particular node, considering its position within the causal graph and the required time frame for the decision to have an effect.
This is particularly relevant, as certain decisions can have an immediate impact, while others may require a longer implementation period. 

\section{Setting the Stage}\label{sec:setting}
\textbf{\textit{Causality}}. 
Given a set $\mathbf{X}$ of $n$ random variables $X_1,\ldots,X_n$,
a Structural Causal Model (SCM) is a tuple $(\mathbf{F},p_\mathrm{U})$ where $\mathbf{F} = \{X_i := f_i(\mathbf{PA}_i, U_i)\}_{i=1}^{n}$ is a set of $n$ structural equations and $p_\mathrm{U}(U_1,\ldots, U_n)$ is a joint distribution over the noise variables $\{U_i\}_1^n$. $f_i$ are deterministic functions computing each variable $X_i$ from its causal parents $\mathbf{PA}_i \subseteq  \mathbf{X} \setminus \{X_i\} $ and its noise variable $U_i$.

In each SCM, the variables within the system are partitioned into two sets: the \textit{exogenous} (unobserved) variables denoted by $\mathbf{U}$ and the \textit{endogenous} (observed) variables denoted by $\textbf{X}$. Endogenous variables are those whose values are influenced by other variables within the system, while exogenous variables are determined by factors outside of the model~\cite{bareinboim2022pearl}.
Besides, an SCM induces a causal graph $\mathcal{G} = \{N, E\}$ where $N = \{N_1, \dots, N_n\}$ is the set of nodes for which $N_i$ represents $X_i$, while $E$ is the set of the edges $E_{ij}$ where $E_{ij} \in E \iff X_i \in \mathbf{PA}_j$. Moreover, it induces an observational distribution over $\mathbf{X}$ to describe what is passively seen or measured, it can also generate many interventional distributions to describe active external manipulation or experimentation. Furthermore, it provides counterfactual statements about what would or could have been, given that something else was observed. These three modes of reasoning are referred to as the three layers of the \quotes{ladder of causation}~\cite{pearl2018book}.
\vspace{2mm}\linebreak
\textbf{\textit{Action Cost}}. 
Identifying the optimal action in the causal AR problem necessarily requires defining a notion of \quotes{intervention cost}, typically using a function 
$ c : \mathbf{X} \times \mathbb{A} \to \mathbb{R}^+$ where $X \in \mathbf{X}$ is the individual. 

We use the notation $\mathbb{A}_\delta \in \mathbb{A}$ to denote an action that changes $X$ by an amount $\delta$. The cost associated with actions that result in greater changes to $X$ is intuitively expected to be higher. In other words, we expect the function $c$ to increase in $|\delta|$ monotonically. The choice of $c$ determines the optimization outcome, regardless of whether the problem is formulated at the observational, interventional, or counterfactual level.
The most widely used cost function in the literature is the $\ell_p$ norm \cite{karimi2020survey}, defined as 
$$c_{\ell_p}(X,\mathbb{A}_\delta)=||\delta||_{\ell_p}=\sqrt[p]{\sum_{i=1}^n |\delta_i|^p}.$$
$c_{\ell_p}$ is often replaced by its normalised variant
$$c_{\ell_pn}(X,\mathbb{A}_\delta)=\sqrt[p]{\sum_{i=1}^n \left(\frac{|\delta_i|}{\sigma_i}\right)^p}, \quad \text{where} \quad \sigma^2_i = Var(X_i),$$
to guarantee scale invariance on the features of $X$.

Despite the constraints inherent in this formulation, it is widely considered a rational and viable choice, mainly due to the inherent challenges of formulating an effective cost function without access to supplementary information.
\\\\
\textbf{\textit{Actionable Recourse}}. The problem of AR can be formulated as a constrained optimization in the following terms: given a binary classification model $h: \mathbf{X} \to \{0,1\}$, and a specific instance $X$ for which $h(X)=0$, the aim is to identify the action $\mathbb{A}_{\delta^*}$ satisfying
$$\delta^* = \left[ \argmin_{\delta}c(X,\mathbb{A}_{\delta})\quad s.t. \quad h(\mathbb{A}_{\delta}(X))=1\right].$$ 
In other words, the objective is to identify the minimal cost action that alters the decision of the model from unfavorable to favorable.
\\
The distinction between AR and the \quotes{\textit{causality-aware}} variant is defined by the manner in which the action $\mathbb{A}_\delta$ operates on a particular instance $X$. In the former $\mathbb{A}_\delta(X):=X+\delta$, whereas in the latter, the action is considered as a causal intervention
$$\mathbb{A}_\delta(X):=\mathbf{F}_{\mathbb{A}_\delta}(X), \quad \text{where} \quad  \mathbf{F}_{\mathbb{A}_\delta}=\{X_i := f_i(\mathbf{PA}_i, U_i)+\delta_i\}_{i=1}^{n}.$$
\section{Motivation}\label{sec:motivation}
In everyday experiences, we typically observe a temporal ordering between the cause and the effect, where the former precedes the latter. This relation could be exemplified by turning on a light switch in a room, where the action of flipping the switch serves as the cause of the light turning on. 
In the context of causal graphs applied to cross-sectional data, time is often ignored, 
leaving room for other notions of dependence between variables.
However, in the framework of AR, it seems natural to include time as a relevant parameter in defining the cost of a specific action. 
We typically assume that \textit{a change in the value of one variable in the causal graph instantaneously affects the descendant variables}. 
In short, probability distributions, including interventional ones, represent a static and unchanging phenomenon of a fundamentally descriptive type. From another perspective, when considering 
a physical system, its structural equations describe the system's behavior
in response to specific physical interventions, ultimately leading to a new and distinct equilibrium state.
However, the propagation of the effects of these interventions to the downstream variables may not occur immediately.
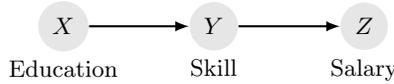
\begin{figure}[t]
    \centering
        \begin{tikzpicture}
        	\node[vertex,label=below:Education] (x) at (-2,0) {$X$};
        	\node[vertex,label=below:Skill] (z) at (0,0) {$Y$};
        	\node[vertex,label=below:Salary] (y) at (2,0) {$Z$};
        	
        	\draw[edge] (x) -- (z);
        	\draw[edge] (z) -- (y);
        \end{tikzpicture} 
    \caption{A causal graph illustrating the relationship between college education, individual skill, and job salary discussed in~\cite{glymour2016causal}.}
    \label{fig:causal_dag}
\end{figure}
For example, Figure~\ref{fig:causal_dag} reports a causal graph consisting of three variables, $X$, $Y$, and $Z$, representing a person's college education, skill, and job salary, respectively. 
We can assume that the system is described by a linear model with additive noise, 
which can be expressed by the following structural equations:
$$ X:=U_X, \quad Y:=a X+U_Y,\quad Z:=b Y+U_Z.$$ 
where, $U_X$, $U_Y$, and $U_Z$ represent noise terms, and $a$ and $b$ are constants. 
Such a model implies that a college education generally leads to better preparation and a higher salary. 
However, the process leading from $X$ to $Y$ can take time (in this case, years), but the model lacks this kind of temporal information 
and thereby is incapable of considering it. 
Suppose a person 
wants to increase his salary $Z$ and queries the model for advice on achieving this goal. 
The alternatives are only two: the person can improve their skills $Y$ by taking a training course, or they can attend college $X$ to obtain skills as a result of the action. 
The optimal action would likely be to take a training course. 
However, the model may not be able to detect this fact. 
In particular, if the coefficient $a$ that links $X$ to $Y$ is sufficiently large, according to~\cite{karimi2020algorithmic}, the optimal action would be to intervene on $X$, rather than on $Y$. 
Generally, whenever a node has many children and/or is the beginning of a long chain, it is likely to be part of the intervention set i.e., the set of variables to intervene on. 
This suggests that the actual formulation of the causal algorithmic recourse problem could be biased towards \emph{root nodes}.

\section{The Role of Time}\label{sec:role_time}
A possible approach to consider the temporal aspect as a relevant parameter of the process is to use the topological information of the graph as an indicator of the time interval between the action and its consequences.
One way to integrate the model with the missing information is to manually assign additional weights to each edge of the graph, incorporating information about the \textit{characteristic response time} of the child variable w.r.t. the change of the parent variable.
This allows for analyzing the dynamic flow of the causal effect over time.

Given $\mathcal{G} = \{N, E\}$, we reformulate the cost function of an action $\mathbb{A}_\delta$ over $X$ \textit{to obtain an effect over $Y$}, as 
$$ c(X,\mathbb{A}_{\delta},Y) = c_{s}\left(X,\mathbb{A}_{\delta}\right) + \lambda c_{t(\mathcal{G})}\left(S(\mathbb{A}_{\delta}),Y\right),$$
where $S(\mathbb{A}_{\delta}) \subseteq N$ is the support of $\mathbb{A}_{\delta}$, i.e. the set of variables \textbf{directly} modified by $\mathbb{A}_{\delta}$. 
We add the parameter $\lambda$ as a free variable to explicitly adjust the balance between the two components of the cost function, namely $c_s$ that denotes the cost function in the feature space and $c_{t(\mathcal{G})}$ that reflects the time part.
It should be noted that the parameters of $c_s$ represent 
features and their values, while $c_{t(\mathcal{G})}$ involves the topological properties of the graph and thus depends on nodes and their graph relationships. In the following, we provide an example of a cost function considering the temporal dimension:
$$ c(X,\mathbb{A}_{\delta},Y) = ||\delta|| + \lambda \sup_{V\in \mathcal{S}(\mathbb{A}_{\delta})} d_{lp}^\mathcal{G}(V,Y),\qquad ||\delta||=\sqrt{\sum_{V\in \mathcal{S}(\mathbb{A}_{\delta})}\delta_V^2},$$
where $d_{lp}^\mathcal{G}(V,Y)$ 
denotes the longest path distance between node $V$ and $Y$ over $\mathcal{G}$.

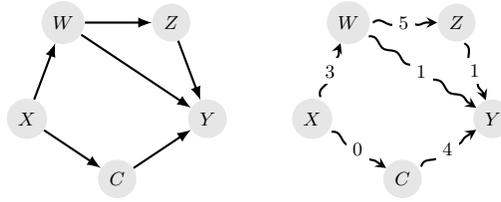
\begin{figure}[!t]
\begin{center}
\begin{tikzpicture}[scale=0.8, transform shape]
    \node[vertex] (x) at (0,-0.3) {$X$};
    \node[vertex] (w) at (0.6,1.3) {$W$};
    \node[vertex] (z) at (2.4,1.3) {$Z$};
    \node[vertex] (c) at (1.5,-1.3) {$C$};
    \node[vertex] (y) at (3,-0.3) {$Y$};
    
    \draw[edge] (x) -- (w);
    \draw[edge] (w) -- (z);
    \draw[edge] (w) -- (y);
    \draw[edge] (z) -- (y);
    \draw[edge] (x) -- (c);
    \draw[edge] (c) -- (y);
\end{tikzpicture}
\hspace{2em}
\begin{tikzpicture}[scale=0.8, transform shape]
    \node[vertex] (x) at (0,-0.3) {$X$};
    \node[vertex] (w) at (0.6,1.3) {$W$};
    \node[vertex] (z) at (2.4,1.3) {$Z$};
    \node[vertex] (c) at (1.5,-1.3) {$C$};
    \node[vertex] (y) at (3,-0.3) {$Y$};
    
    \draw[snakedge] (x) -- (w)
    node[scale=1,midway,fill=white] {$3$};
    \draw[snakedge] (w) -- (z)
    node[scale=1,midway,fill=white] {$5$};
    \draw[snakedge] (w) -- (y)
    node[scale=1,midway,fill=white] {$1$};
    \draw[snakedge] (z) -- (y)
    node[scale=1,midway,fill=white] {$1$};
    \draw[snakedge] (x) -- (c)
    node[scale=1,midway,fill=white] {$0$};
    \draw[snakedge] (c) -- (y)
    node[scale=1,midway,fill=white] {$4$};
\end{tikzpicture}
\end{center}
\caption{An example of a causal graph and a simple way to incorporate time information over it. A wavy edge weight $\tau_{ij}$ is meant to represent time between intervention over node $i$ and the observed effect on node $j$.}\label{fig:graphexample}
\end{figure}

Figure \ref{fig:graphexample} illustrates the graphical example of the above formula. 
Assuming for simplicity that each edge has 
a weight of $1$, we consider an action over the set $\{W,X\}$, observing different directed paths towards $Y$. 
By applying $c$ and focusing on the \emph{time} part of the formula, we obtain $$c_t = \sup_{V\in \{X,W\}}d_{lp}(V,Y) = d_{lp}(X,Y) = 3.$$ 
The $sup$ operation is motivated by 
the following observation: 
when we act on some variables and interpret the process as a dynamic information flux from them to the target, the total causal effect will be observed after every single process finishes, equivalently, after the last one does.
This proposal for $c_t$ may be overly simplistic and lacking flexibility, as adding a single edge with a high weight can radically change the cost function regardless of its causal effect. 
A less rigid formulation of $c_t$ could consider the average response time across the multitude of causal processes involved, each of which can be represented by a causal path between the intervention variable and the target. 
The average could be weighted by the causal impact of each process, ensuring that minor processes do not excessively skew the evaluation.

Consider the case of linear models with additive noise characterized by SCMs of the following form:
$$ X_j := \sum_{X_i \in \mathbf{PA_j}} \beta_{ij} X_i + U_j.$$
In this setting, the path impact can be assessed by calculating the product of the coefficients $\beta_{ij}$ associated with its edges. 
Furthermore, the total causal effect of a cause $X$ over $Y$ is the sum of the effects attributed to the paths between $X$ and $Y$.
Suppose we also have information regarding the response times associated with edges in $\mathcal{G}$, as depicted in Figure \ref{fig:timegraph}. We denote these quantities as $\tau_{ij}$.
Considering the set of all these paths and defining the weight of each path $\pi$ as $w_\pi := \prod_{E_{ij} \in \pi}\beta_{ij}$, we define the following:
$$ c_t(X,Y) = \frac{1}{Z_{XY}}\sum_{\pi | X\overset{\pi}{\to }Y}w_\pi t_\pi, \quad Z_{XY} := \sum_{\pi | X\overset{\pi}{\to }Y}w_\pi, \quad t_\pi := \sum_{E_{ij} \in \pi}\tau_{ij}.$$
In brief, $t_\pi$ represents the propagation time of the causal effect \textit{through} $\pi$, and $c_t$ is the weighted average of $t_\pi$, weighted by the relative importance of the causal effect of each path on the overall process. 
The value of $c_t$ in this revised formulation intuitively represents the time at which a significant portion of the causal effect becomes observable, rather than necessarily capturing its entirety. 
In this sense, it offers greater robustness and flexibility compared to a cost that relies solely on calculating the longest path.
\\\\
\textbf{\textit{General Remarks}}. 
We have presented some formulations to incorporate temporal dimension within the causal AR framework. 
However, it is essential to recognize that a general solution is insufficient in resolving the issue across diverse contexts. Instead, deeper considerations must be given to the unique demands of the user and the specific properties inherent in the problem being addressed. 

If the user needs to complete the action within a time constraint, e.g., purchase a house and move in within two months of applying for a loan, instead of directly adding the term $c_t$ to the total cost, an alternative approach could be to use it as a constraint in the optimization problem. This approach would redefine the problem as finding the most cost-effective action, where the effects are achieved within a predetermined maximum time, leading to a favorable decision.

Furthermore, the problem could be such that $\tau_{ij}$ is not a predetermined fixed value, but rather it may depend on the specific instance $x$ under consideration. For instance, in the context of achieving fitness through a diet and exercise program, the required time would vary based on factors such as the user's age, gender, and current weight.
In such cases, it would be necessary to enrich the data structure and fit $\tau$ based on available data. 

Lastly, the nature of the data can have a significant impact on the formulation of the problem. As an example, where data provides relevant information regarding the temporal dimension, the choice of $c_t$ may inherently depend on it. In particular, when working with a time series dataset, it may be possible to construct a more nuanced $c_t$ that aligns with specific practical contexts.

Regarding the aforementioned consideration, this issue is complex and multifaceted, presenting an intriguing and fruitful research area. While it is beyond the scope of this work to fully tackle it, the development of effective methods necessitates applying our proposal to real-world problems. 


\section{Experiments}\label{sec:experiment}
The experimental evaluation\footnote{The code is available here: \url{https://github.com/marti5ini/time-car/}} aims to show that the current formulation of causal AR may have a bias towards root nodes as postulated in Section \ref{sec:motivation}, thus highlighting the significance of the role of time in this field.
\vspace{-10pt}
\subsection*{Experimental Setup}\label{sec:exp_setup}
\begin{figure}[!t]
\centering
\begin{tikzpicture}[scale=1, transform shape]
    \node[vertex] (g) at (-2,0) {$G$};
    \node[vertex] (a) at (0,2) {$A$};
    \node[vertex] (e) at (0,0) {$E$};
    \node[vertex] (j) at (2,0) {$J$};
    \node[vertex] (l) at (2,2) {$L$};
    \node[vertex] (d) at (4,0) {$D$};
    \node[vertex] (i) at (4,2) {$I$};
    \node[vertex] (s) at (6,0) {$S$};
    \node[vertex,draw=Cerulean, very thick] (y) at (6,2) {$Y$};
    
    \draw[edge] (g) -- (e);
    \draw[edge] (a) -- (e);
    \draw[edge] (g) -- (l);
    \draw[edge] (a) -- (l);
    \draw[edge] (e) -- (j);
    \draw[edge] (a) -- (j);
    \draw[edge] (g) to [out=-30,in=210] (j);
    \draw[edge] (g) to [out=-30,in=210] (d);
    \draw[edge] (a) -- (d);
    \draw[edge] (l) -- (d);
    \draw[edge] (j) -- (i);
    \draw[edge] (g) -- (i);
    \draw[edge] (e) -- (i);
    \draw[edge] (i) -- (s);
    \draw[edge] (l) to [out=30,in=150] (y);
    \draw[edge] (d) -- (y);
    \draw[edge] (i) -- (y);
    \draw[edge] (s) -- (y);
\end{tikzpicture}
\caption{The \textit{German Credit} inspired causal DAG.}
\label{fig:german}
\end{figure}
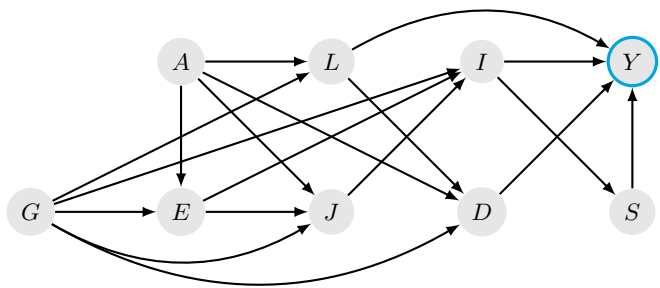

We consider a semi-synthetic SCM based on the German Credit dataset\footnote{\url{https://www.kaggle.com/datasets/uciml/german-credit}}. While the corresponding causal graph is shown in Figure~\ref{fig:german}, the loan approval SCM consists of the following structural equations and noise distributions: 
\begin{center}
    \resizebox{0.95\textwidth}{!}{%
        \begin{minipage}{\textwidth}
            \begin{align*}
            (Gender) \qquad G &:=U_G, \qquad &U_G \sim \operatorname{Bernoulli}(0.5) \\[0.3em]
            (Age) \qquad A &:=-35+U_A, \qquad &U_A \sim \operatorname{Gamma}(10,3.5)\\[0.3em]
            (Education) \qquad E &:= G+ A +U_E, \quad &U_E \sim \mathcal{N}(0,1)\\[0.3em]
            (Job) \qquad J &:= G + 2 A + 4 E + U_J, \quad &U_J \sim \mathcal{N}(0,2)\\[0.3em]
            (Loan\; Amount) \qquad L&:= A + 0.5 G+ U_L, \quad &U_L \sim \mathcal{N}(0,3)\\[0.3em]
            (Loan\; Duration) \qquad D&:=G - 0.5 A + 2L+U_D, \quad &U_D \sim \mathcal{N}(0,2)\\[0.3em]
            (Income) \qquad I&:=0.5G + A + 4 E + 5 J + U_I, \quad &U_I \sim \mathcal{N}(0,4)\\[0.3em]
            (Savings) \qquad S&:= 5 I +U_S \quad &U_S \sim \mathcal{N}(0,2)
            \end{align*}
         \end{minipage}
    }
\end{center}
The target $Y$ is obtained according to:
$$
Y \sim \sigma (2I+3S-L-D) .
$$

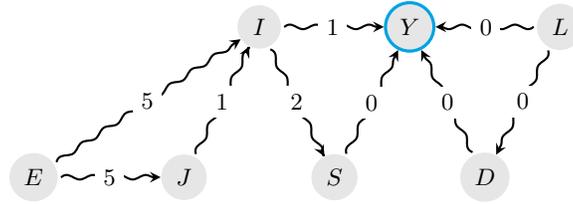
\begin{figure}[!t]
\centering
\begin{tikzpicture}
    \node[vertex] (e) at (-2,0) {$E$};
    \node[vertex] (l) at (5,2) {$L$};
    \node[vertex] (j) at (0,0) {$J$};
    \node[vertex] (d) at (4,0) {$D$};
    \node[vertex] (i) at (1,2) {$I$};
    \node[vertex] (s) at (2,0) {$S$};
    \node[vertex,draw=Cerulean, very thick] (y) at (3,2) {$Y$};
    \draw[snakedge] (e) -- (i)
    node[scale=1,midway,fill=white] {$5$};
    \draw[snakedge] (e) -- (j)
    node[scale=1,midway,fill=white] {$5$};
    \draw[snakedge] (j) -- (i)
    node[scale=1,midway,fill=white] {$1$};
    \draw[snakedge] (i) -- (s)
    node[scale=1,midway,fill=white] {$2$};
    \draw[snakedge] (d) -- (y)
    node[scale=1,midway,fill=white] {$0$};
    \draw[snakedge] (i) -- (y)
    node[scale=1,midway,fill=white] {$1$};
    \draw[snakedge] (s) -- (y)
    node[scale=1,midway,fill=white] {$0$};
    \draw[snakedge] (l) -- (y)
    node[scale=1,midway,fill=white] {$0$};
    \draw[snakedge] (l) -- (d)
    node[scale=1,midway,fill=white] {$0$};
\end{tikzpicture}
\caption{Actionable weighted DAG where the coefficients represent the response times for parent-child relationships.}
\label{fig:timegraph}
\end{figure}

We model \textit{Age}, \textit{Gender} and \textit{Loan Duration} as non-actionable variables but consider the latter to be mutable, i.e., it cannot be manipulated directly but is allowed to change (e.g., as a consequence of an intervention on \textit{Loan Amount}).
Our analysis specifically focuses on \textit{Education} and \textit{Income}, as we suppose that these two variables may display the most significant bias. The reason behind our belief is that both variables lie on the same causal pathway. Moreover, from a semantic perspective, one would anticipate that education is a way slower process compared to an income increase. A potential illustration of this idea is depicted in Figure \ref{fig:timegraph}, in which edges starting from \textit{Education} have quite significant weights compared to the others.

\begin{figure}[!t]
\centering
\includegraphics[width=6cm]{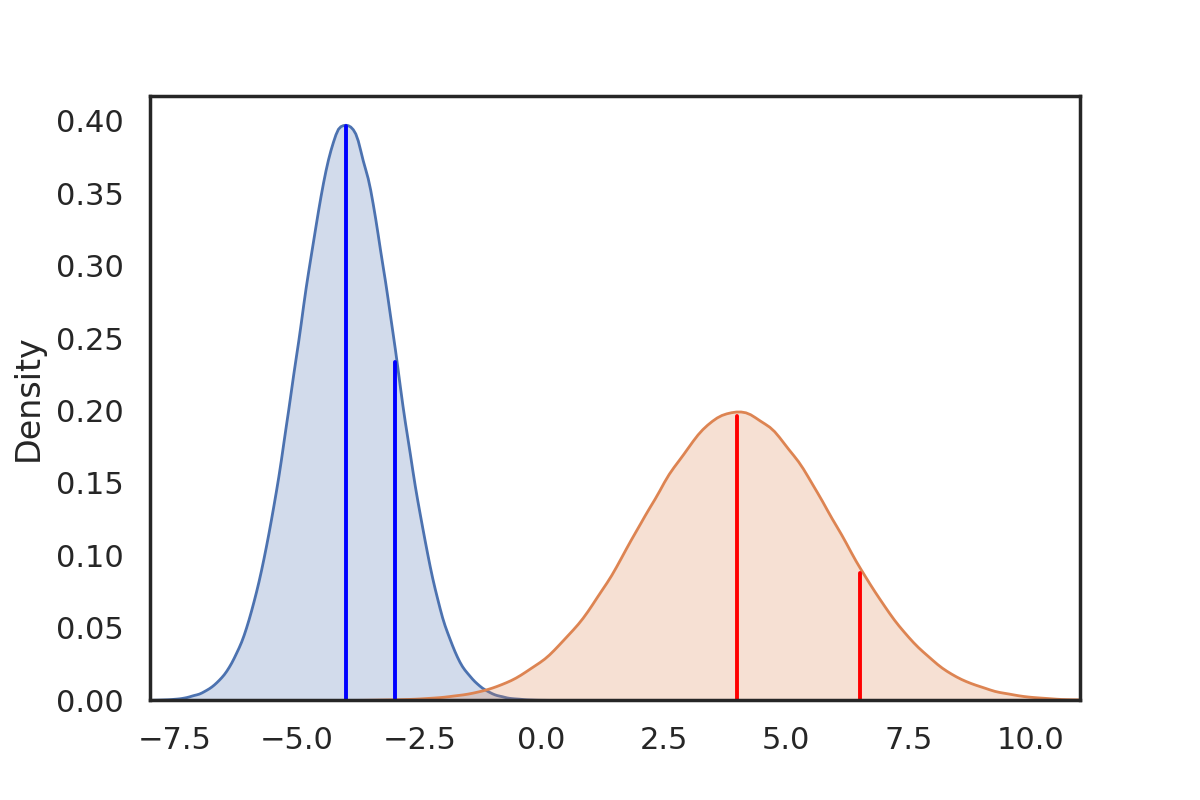}
\caption{Example of possibly different costs based on the scale of distributions considered.}
\label{fig:costo}
\end{figure}

Notably, the structural equations used in \cite{karimi2020algorithmic} differ from those defined in our work. 
The rationale for the variation is rooted in the observation that the SCM used had excessive noise to each variable, thus rendering the causal effect of any reasonable intervention practically irrelevant and precluding the possibility of testing alternative scenarios. 
Moreover, while the original structural equations accounted for both linear and nonlinear relationships, it was deemed sufficient for us to only consider the linear ones. 
Indeed, for the scope of our work, studying systems other than linear was unnecessary because the time dimension is independent of the SCM form.
\vspace{-5pt}
\\\\
\textbf{\textit{Proper Variance}}.
As stated in Section \ref{sec:setting}, the cost function used in \cite{karimi2020algorithmic} is the normalized $\ell_1$ norm. 
Such normalization is required because if the distributions of the variables being manipulated have different scales, this could result in different costs. Figure \ref{fig:costo} depicts a possible scenario that describes the aforementioned phenomenon. The blue-colored distribution has a smaller $\sigma$, while the red has a larger one. 
As a result, the intervention cost in the case of the first distribution is higher than that of the second, given the same amount of $\delta$.

Although this formulation is commonly used, we believe it may not be sufficient, given that it entails critical aspects that we endeavor to explain below.

For instance, we consider a simple SCM system, such as the linear Gaussian model. 
In this system, each equation can be defined by the following formula:
$$ X_i := \sum_{X_j \in \mathbf{PA_i}} a_{ji} X_j + U_i,\quad U_i \sim \mathcal{N}(0,\hat{\sigma}^2_i), \quad a_{ji} \in \mathbb{R}.$$
$\hat{\sigma}^2_i$ \textit{can be interpreted as the variability of $X_i$ due to the exogenous variables of the system.} Specifically,  we consider it as the \textit{proper variance} of $X_i$. Such variance is useful to observe that $\hat{\sigma}^2_i$ and $\sigma^2_i=Var(X_i)$ are not the same quantity since the latter inherits the variability of the parents of $X_i$. 
 An approximation of the magnitude of $\sigma^2_i$ - assuming for simplicity independence among the components $X_j$ - can be obtained using the formula
$$\sigma^2_i=Var(X_i) \approx \sum_{X_j \in \mathbf{PA_i}} a_{ji} \sigma^2_j + \hat{\sigma}^2_i,$$
We observe that $\sigma_i\geq \hat{\sigma}_i$ holds. 
Furthermore, if $X_i$ is an ancestor of $X_j$, the structural equation of $X_j$ can be rewritten as a regression containing the term $a_{ij} X_i$ and other terms. 
If $a_{ij}\geq 1$, we have a \emph{avalanche effect} of the variances. This means that the variables become increasingly spread along the causal order. A comprehensive discussion on this topic can be found in \cite{reisach2021beware,reisach2023}, where this property is exploited to infer the causal structure of a directed acyclic graph.

When considering AR, we argue that using the normalized $\ell_1$ norm in relation to ${\sigma}^2_i$ as a cost function can lead to significant distortions if applied to the system being studied. This is because interventions become less expensive as one moves down the graph's topological order, regardless of each variable's internal properties. To address this issue, \textit{we have decided to normalize the cost function based on the proper variances} $\hat{\sigma}^2_i$.
\\\\
\textbf{\textit{Dataset}}. Using the loan approval SCM described above, we have generated a synthetic observational dataset $\mathcal{D}$ with $10000$ samples. For each feature $X_i$, we applied an intervention described by: 
$$X_i \rightarrow X_i + \alpha \hat{\sigma}_{i}, \quad \text{where} \quad \alpha \in \mathbb{R}^+, $$ 
generating for each intervention an interventional dataset $\mathcal{D}_{X_i}$\footnote{Note that the cost of the action is equal to $\frac{\alpha\hat{\sigma}_i}{\hat{\sigma}_i}=\alpha$, regardless of the variable on which the action is executed. Given the same cost, we are interested in determining which action has the greatest impact on the target variable $Y$.}.
To estimate the \quotes{derivative} of the causal effect, we have utilized the following formula:
\begin{equation}\label{eq:estimate}
\frac{\overline{Y}_{\mathcal{D}_{X_i}} - \overline{Y}_{\mathcal{D}}}{\alpha}.
\end{equation}

\subsection{Results}
In Table \ref{tab:causal-effect-derivative-scm}, we report each feature's estimated Causal Effects Derivative (\texttt{CED}) obtained by applying Equation \ref{eq:estimate} respectively to the SCM reported in~\cite{karimi2020algorithmic} and the one proposed in this work. With regard to the former, it can be observed that nearly every \texttt{CED} is close to zero, giving rise to an unrealistic problem. 
\begin{wraptable}{r}{0.5\textwidth}
\vspace{-25pt}
\footnotesize
\centering
\caption{Comparison between the estimated Causal Effect Derivative (\texttt{CED}) of the SCM used in \cite{karimi2020algorithmic} and the ones we proposed. Non-actionable variables are in red, best estimated actionable values are in bold.}
\vspace{10pt}
\begin{tabular}{|c|c|c|}
\hline
\textbf{Feature} & \textbf{\texttt{CED} in \cite{karimi2020algorithmic}} & \textbf{Our \texttt{CED}} \\ \hline
\textcolor{red}{Gender} & \textcolor{red}{0.004} & \textcolor{red}{0.273} \\ 
\textcolor{red}{Age} & \textcolor{red}{-0.016} & \textcolor{red}{0.329} \\ 
Education & 0.000 & \textbf{0.181} \\ 
Loan Amount & -0.018 & -0.099 \\ 
Job & 0.015 & 0.087 \\ 
\textcolor{red}{Duration} & \textcolor{red}{-0.021} & \textcolor{red}{-0.037} \\ 
Income & \textbf{0.058} & 0.137 \\ 
Savings & 0.038 & 0.066 \\\hline
\end{tabular}
\label{tab:causal-effect-derivative-scm}
\vspace{-25pt}
\end{wraptable}
For instance, the absence of any importance of variables such as \textit{Gender}, \textit{Age} and \textit{Education} in the context of credit score prediction appears highly unusual. One possible explanation could be related to the previous discussion on the use of variance as a normalization coefficient. If the variance increases along the causal direction (multiplying at each edge), variables that are further away from the target will have a smaller variance and, therefore, a higher intervention cost. Conversely, those closer to the target will have a larger variance and a lower cost. In fact, the four variables with the highest \texttt{CED} are at a distance of $1$ from Y, while all those at a distance of at least $2$ have lower \texttt{CED} values.
\begin{figure}[h!]
\centering
\includegraphics[width=1\textwidth]{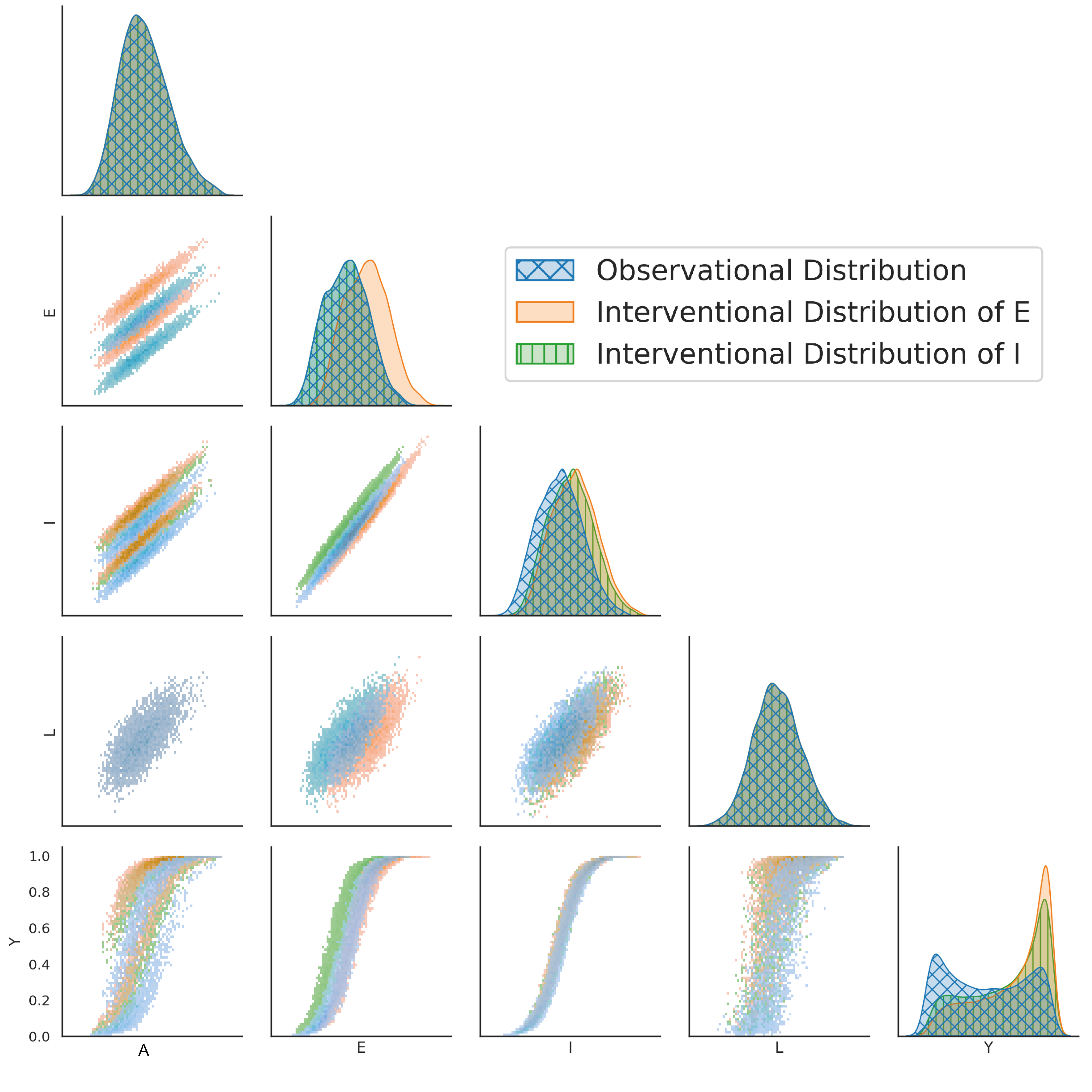}
\caption{Pair plot representing observational and interventional distributions of SCM described in Section~\ref{sec:exp_setup}. 
The results indicate that intervening on \textit{Education} is more effective than intervening on \textit{Income} when it comes to improving the outcome of $Y$, as depicted in the bottom-right subplot.}
\label{fig:plot}
\end{figure}
\vspace{-10pt}

Figure \ref{fig:plot} presents the pair plot of the SCM system described in Section~\ref{sec:exp_setup}, for three different distributions: Observational, Interventional on \textit{Education}, and Interventional on \textit{Income}. 
Due to space constraints, only a subset of variables was selected, focusing on the two treatment variables and on $Y$. We added \textit{Age} and \textit{Loan Amount} to provide a clearer view of the system's structure and how interventions modify instances. Regarding \textit{Loan Amount}, it is observed that since it does not depend on either \textit{Education} or\textit{ Income}, its distribution remains the same in all cases. The same applies to \textit{Age}, which provides additional insight: the pair plots (\textit{Age, Education}) and (\textit{Age, Income}) show a bimodal distribution, explainable through the presence of fork paths $A\rightarrow E \leftarrow G$ and $ A\rightarrow E\rightarrow I\leftarrow G$, revealing the presence of a gender gap in \textit{Education} and \textit{Income} of individuals described by the SCM. 
Finally, the most significant observation pertains to the \textit{Target} variable $Y$. Notably, it is evident that for the same cost, \textit{Education} has a stronger causal effect than \textit{Income}. As a result, the recommended action for addressing the unfavorable outcome through recourse would involve prioritizing the improvement of \textit{Education}, confirming the bias towards the root nodes we hypothesized in Section \ref{sec:motivation}.

\section{Discussion}\label{sec:findings}
\textbf{\textit{Broader Impact}}.
A crucial aspect emphasized by the XAI community is the user's demand for plausible explanations~\cite{guidotti2019survey,ustun2019}. In the context of AR, \textit{plausibility} refers to the perceived consistency and reasonableness of the recommendations provided by recourse approaches. 
From a psychological perspective, providing plausible explanations enables users to form mental models that align with their prior knowledge and reasoning abilities \cite{beretta2022}. 
When the temporal dimension is incorporated into causal reasoning, an AR approach could ensure that the actions suggested are psychologically congruent with human intuitions and mental frameworks. This compatibility fosters a sense of trust and confidence in the algorithmic system, thereby facilitating user acceptance and engagement. 

Furthermore, \textit{actionability} is considered one of the crucial aspects in a counterfactual generation process, as highlighted in \cite{guidotti2022counterfactual}. We propose expanding the concept beyond the notion of \textit{being able to act upon} to include the ability to do so within a reasonable timeframe. In fact, if the action required to implement a recommendation is excessively time-consuming or impractical, the recommendation becomes unhelpful for the user. 

In conclusion, the findings of this study could have significant implications for the XAI field. 
By incorporating the temporal dimension into causal AR reasoning, the plausibility of explanations is enhanced, aligning them more closely with user expectations and cognitive processes. Moreover, the consideration of actionability addresses the user's need for practical and timely actions. 
These insights highlight that time-aware causal recourse approaches are a valuable advancement, as they bridge the gap between human intuition, psychological congruence, and efficient decision-making processes.
\vspace{-5pt}
\\\\
\textbf{\textit{Prediction vs Improvements}}.
Another point necessary to clarify is the relationship between the predicted outcome ($\hat{Y}$) and the actual outcome ($Y$).  
ML models are typically statistical in nature and do not inherently capture causal relations.
$\hat{Y}$ may behave very differently from $Y$. 
For example, consider the diagnosis of a disease in the medical field. Suppose we have several variables that indicate the presence or absence of certain symptoms in a patient. The predicted outcome, $\hat{Y}$, will utilize the correlation between these symptoms and the disease in order to enhance the predictive capability of the model.
If the patient were to take a drug capable of suppressing some of these symptoms, $\hat{Y}$ could change significantly, regardless of whether the drug is effective in curing the underlying disease. 
However, the presence or absence of the disease $Y$ would not change at all. 
From a causal viewpoint, taking action on effects does not have any impact on causes. For a more comprehensive analysis of this aspect, we recommend referring to \cite{konig2022}.
In our work, we defined a predictive model aware of the causal relationships between variables, so that $\hat{Y}$ aligns with $Y$, allowing us to treat them as a unified entity. 
From this perspective, it is worth noting that the outcome is not directly influenced by all the variables within the system, but rather by a specific subset of them. Regarding \textit{Gender, Age, Education, Job}, their impact on the outcome $Y$ is only indirect and mediated through other variables. A purely statistical and non-causal model would ignore entirely these variables, as it would observe, for instance, that $\hat{Y}$ is independent of $J$ given $I$.

\section{Conclusions}\label{sec:conclusions}

In this work, we discussed the problem of Algorithmic Recourse from a causal perspective, focusing on incorporating the temporal dimension into the evaluation of the cost of an action. Firstly, we justified this integration methodologically by discussing its connection with the causal graph's topological structure and proposing a new \textit{time-aware} causal AR formulation. Then, we tested our theoretical intuition through an experiment inspired by the credit score model on the German Credit Dataset, confirming our expectations: if the causal model is unaware of the response times between variables, it could recommend actions that, although optimal considering only the SCMs, would require too much time to be practically viable in most applications. These results serve as a strong motivation for future work to develop and evaluate causal algorithms that effectively incorporate temporal information to enhance the quality of recommendations.

As a final remark, we would like to point out how incorporating the temporal dimension into AR is a conceptually distinct problem from its causal formulation. A very similar discussion to the one presented in this work could be made under different causal knowledge conditions, up to the absence of it. In our opinion, the advantage of the causal framework stems from the use of the graph structure as a support for the finer estimation of the temporal relationships between the system's variables, requiring at least knowledge of the causal graph.

\bigskip
\textbf{Acknowledgments.}
Work supported by the European Union’s Horizon 2020 research and innovation programme under the Excellent Science European Research Council (ERC) programme for the XAI project (g.a. No. 834756), and by the FAIR (Future Artificial Intelligence Research) project, funded by the NextGenerationEU program within the PNRR-PE-AI scheme (M4C2, investment 1.3, line on Artificial Intelligence). This work reflects only the authors’ views and the European Research Executive Agency (REA) is not responsible for any use that may be made of the information it contains.

\newpage

\bibliographystyle{splncs04}
\bibliography{biblio}
\end{document}